\documentclass{article}
\usepackage{spconf,amsmath,graphicx}
\usepackage{hyperref}
\usepackage{multirow,multicol, makecell}
\usepackage{multirow}
\usepackage{multicol}
\usepackage{bm}
\usepackage{amssymb}
\usepackage{booktabs}
\usepackage{enumitem}
\usepackage{amsthm,amsmath,amssymb}
\usepackage{tabularx}
\usepackage{mathrsfs}
\usepackage{booktabs}

\title{CONCSS: Contrastive-based Context Comprehension for Dialogue-appropriate Prosody in Conversational Speech Synthesis}

\name{
    Yayue Deng$^1$, Jinlong Xue$^1$, Yukang Jia$^3$, Qifei Li$^1$, Yichen Han$^1$ \\
    Fengping Wang$^1$, Yingming Gao$^1$, Dengfeng Ke$^2$, Ya Li$^{1,*}$\thanks{* Ya Li is the corresponding author.}
}

\address{
  $^1$Beijing University of Posts and Telecommunications, Beijing, China\\
  $^2$Beijing Language and Culture University, Beijing, China\\
  $^3$Perfect World Co., Ltd, Beijing, China
}

\begin{document}
\ninept
\maketitle
\begin{abstract}
Conversational speech synthesis (CSS) incorporates historical dialogue as supplementary information with the aim of generating speech that has dialogue-appropriate prosody. While previous methods have already delved into enhancing context comprehension, context representation still lacks effective representation capabilities and context-sensitive discriminability. In this paper, we introduce a contrastive learning-based CSS framework, CONCSS. Within this framework, we define an innovative pretext task specific to CSS that enables the model to perform self-supervised learning on unlabeled conversational datasets to boost the model's context understanding. Additionally, we introduce a sampling strategy for negative sample augmentation to enhance context vectors' discriminability. This is the first attempt to integrate contrastive learning into CSS. We conduct ablation studies on different contrastive learning strategies and comprehensive experiments in comparison with prior CSS systems. Results demonstrate that the synthesized speech from our proposed method exhibits more contextually appropriate and sensitive prosody.
\end{abstract}

%

\begin{figure*}[ht]
  \centering
  \includegraphics[width=0.93\linewidth]{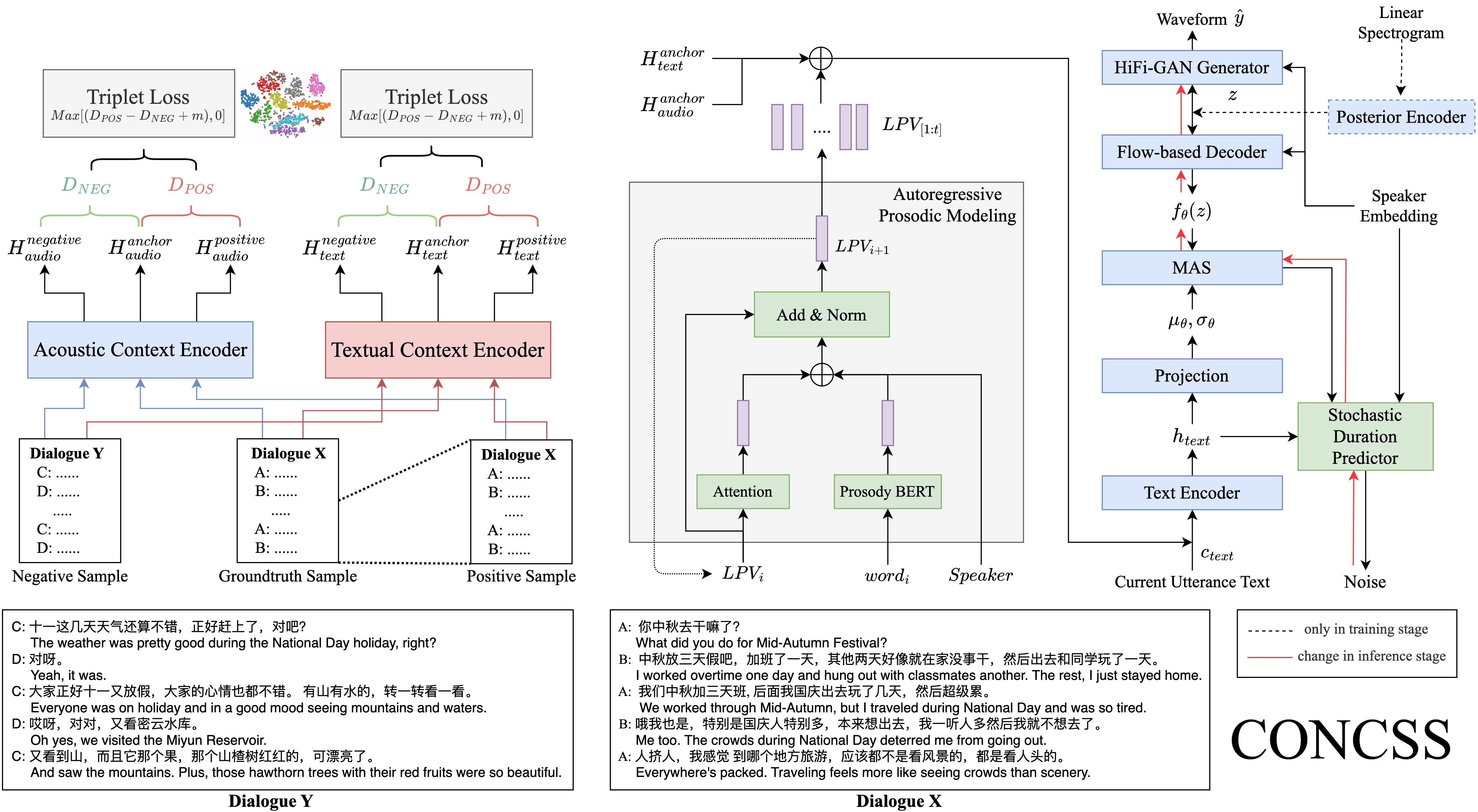}
  \caption{Illustration of our proposed CONtrastive-based Conversational Speech Synthesis (CONCSS).}
  \label{pic:overview}
  \vspace{-0.4cm}
\end{figure*}

\section{Introduction}
\label{sec:intro}
Recent advancements in speech synthesis systems \cite{shen2023naturalspeech,DBLP:conf/icassp/tacotron2,DBLP:conf/iclr/FastSpeech2,DBLP:conf/icml/vits} have enabled the generation of high-quality speech. However, in some complex scenarios, such as human-computer interaction (HCI), these systems still fall short since they are unable to generate audio with natural and human-like prosody. 

Previous psychological studies \cite{stephens2010speaker,lyu2019neural} show that when processing conversation in real-time, our brain quickly uses various information, such as prior statements and the speaker's identity, to help understand the current speech. Similar to human communication, several studies \cite{gallegos2021comparing,lei2022towards,DBLP:conf/interspeech/SridharSCB11,lee2022dailytalk,xue2023M2CTTS} demonstrate that incorporating historical dialogues as supplementary information into speech synthesis can improve the model's understanding of prior statements, thereby helping to enhance the prosody of the synthesized audio. Hence, a greater interest has been shown in developing conversational speech synthesis (CSS) task which focuses on improving the model's context-understanding capability and generating audio with context-appropriate prosody.

Guo et al. \cite{guo2021agent} first introduce a GRU-based context modeling method to extract semantic information from the history context at the utterance level. Moreover, further studies \cite{nishimura2022acoustic,xue2023M2CTTS,DBLP:journals/corr/fctalker} have verified that combining acoustic and textual context can greatly improve the naturalness of synthesized speech. Nishimura et al. \cite{nishimura2022acoustic} utilize cross-modal attention to capture both long-term linguistic and prosodic context information.
Xue et al. \cite{xue2023M2CTTS} and Hu et al. \cite{DBLP:journals/corr/fctalker} combine both fine-grained and coarse-grained context encoders to provide sufficient information for better context comprehension. Meanwhile, some graph-based methods \cite{DBLP:conf/icassp/graphCSS,DBLP:conf/mm/msrgcn} are proposed to capture multi-scale context dependencies between different modalities. 

While prior CSS frameworks have demonstrated the capability to enhance context comprehension, thereby yielding context-related prosody, the question remains: Is this output vector of the context encoder sufficiently indicative of the underlying context variations? This is a concern since these CSS approaches rely solely on jointly training the acoustic model and context encoder using the mel-reconstruction loss. Without explicit constraints, the latent vector derived from the context encoder may not possess desired characteristics like interpretability, strong representative capacity, and context-sensitive discriminability. Hence, it is imperative to include an additional constraint in order to obtain better results for context representation extraction from a jointly-trained model.

Motivated by this, we propose a novel conversational speech synthesis framework CONCSS which combines contrastive learning and CSS to learn effective and context-sensitive representation. To this end, we define a novel pretext task specific to CSS as an important strategy to learn context representations using pseudo labels. By doing so, the context encoder is then forced to learn what we care about, e.g., underlying semantics and discernible context variations. To validate the effectiveness of the method, we conduct comprehensive and uniquely designed experiments. Results demonstrate that our proposed method can enhance discriminability and context sensitivity of context vectors compared with previous methods. Furthermore, the context-sensitive vectors guide downstream acoustic model to synthesize audio with more context-appropriate prosody. Our work has three main contributions:
    \begin{enumerate}[topsep=0pt, partopsep=0pt, itemsep=0pt, parsep=1pt, leftmargin=*]
    \item The primary contribution is the novel CSS framework CONCSS for enhancing prosody, driven by contrastive learning to address the context understanding issue from a self-supervised perspective. To the best of our knowledge, we are the first to adopt contrastive learning for the CSS task. 
    \item To address the issue of context representation in CSS, we design an innovative pretext task specific to CSS tasks, along with a sampling strategy. This approach compels the context encoder to generate distinct representations for diverse scenarios, thereby promoting both context sensitivity and distinctiveness in prosody.
    \item We comprehensively evaluate models on their ability to produce context-sensitive vectors and dialogue-appropriate prosody. Specifically, we compare prosodic performance between different CSS frameworks when they are exposed to diverse contexts. 
    \end{enumerate}
The rest of the paper is organized as follows. Section~\ref{sec:method} introduces our proposed method. Experimental setup and analysis of results are shown in Section~\ref{sec:experiments}. Finally, Section~\ref{sec:conclusions} concludes the paper. 
\vspace{-0.2cm}
\section{PROPOSED METHOD}
\vspace{-0.15cm}
\label{sec:method}
To strengthen context comprehension capability of the context encoder and develop context-sensitive context representation without using labeled data, we propose a novel self-supervised contrastive-based framework where we enhance the prosody of speech from context-related to context-sensitive.
\vspace{-0.4cm}
\subsection{Problem Definition and Framework Overview}
\vspace{-0.1cm}
Assume an input sequence of $N$ utterances $[(u_{1}, p_{1}), (u_{2},p_{2}),....,\\ (u_{N},p_{N})]$ in a conversation where each utterance $u_i$ is spoken by speaker $p_i$. In the CSS task, considering a context length of $i$, the context encoder takes historical utterances $[(u_{1}, p_{1}), (u_{2},p_{2}),....,\\ (u_{i},p_{i})]$ as input and generates context vector $h_i$. Then, context vector $h_i$ is integrated into the downstream speech synthesis framework, generating audio with dialogue-appropriate prosody.

As illustrated in Figure~\ref{pic:overview}, the proposed framework comprises the following four enhancements to the latest speech synthesis model VITS \cite{DBLP:conf/icml/vits}:
1) Leveraging an innovative pretext task, context-dependent pseudo-labels are created to direct the model toward obtaining desired attributes (e.g., context-sensitive discriminability and interpretability); 2) We incorporate an acoustic and textual context encoder, maintaining structure consistent with \cite{xue2023M2CTTS}; 3) We employ triplet loss \cite{DBLP:conf/cvpr/SchroffKP15} coupled with a hard negative sampling strategy to amplify the model's context awareness and comprehension, thereby endowing the synthesized speech with context-sensitive distinctiveness and context-appropriate prosody; 4) We utilize an autoregressive prosodic modeling (APM) module with a pre-trained prosodic language model \cite{bert}.

\subsection{Context-aware Contrastive Learning}
\label{sec:CL}
\subsubsection{Pretask Definition}
We hypothesize that a context encoder with adept context understanding capabilities can detect variations in prior statements and, furthermore, exhibit appropriate and distinguish representations to different contexts. 

Hence, the context-based pretext task is defined as follows: 
Positive sample $h_i^p$ is the output of context encoder given by the same historical dialogue with the groundtruth sample but has different context lengths, whereas negative sample $h_i^n$ is given by different dialogues with non-overlapping context backgrounds.

Thus we want context vectors can satisfy the following criteria:
\vspace{-0.2cm}
\begin{equation}
\begin{split}
    D (h_i,h_i^p) & < D (h_i, h_i^n) \\
\end{split}
\vspace{-0.2cm}
\end{equation}
where similarity measurement $D(\cdot)$ is defined as squared Euclidean distance in the context representation space.
 \vspace{-0.2cm}
\subsubsection{Sampling Strategy}
\label{sec:sample_select}
Furthermore, previous work \cite{DBLP:conf/cvpr/SchroffKP15} has demonstrated that hard negative samples provide the most significant assistance to update the gradients in the optimization process.
In order to select hard negative samples, we consider that negative samples should be derived from two sources: 1) from dialogues with the same speakers but in entirely disparate contexts (intra-speaker classes); 2) from dialogues involving different speakers with non-overlapping context backgrounds (inter-speaker classes).

In our case, intra-speaker classes are hard negative samples because of the potential similarities for context vectors of the same speaker. Although the context variations are totally different, the context spoken by the same speakers may lead to similar speaker-related prosody. Hence, we further employ intra-speaker classes as a hard negative sampling strategy to enhance contrastive learning.
\vspace{-0.2cm}
\subsubsection{Multi-modal Context Comprehension with Triplet Loss}
To achieve context-sensitive discriminability of context vector, we directly maximize the similarity between positive pairs and minimize the similarity of negative pairs via a triplet loss \cite{DBLP:conf/cvpr/SchroffKP15}, which can be expressed as:
\vspace{-0.1cm}
\begin{equation}
     L(h_i^a,h_i^p,h_i^n) = \max \{D(h_i^a,h_i^p) - D(h_i^a,h_i^n) + {\rm m}, 0\}
    \label{equ:1}
    \vspace{-0.1cm}
\end{equation}
where the margin parameter $m$ imposing the distance between negative samples and positive samples should be larger than $m$. The anchor $h_i^a$, positive $h_i^p$, and negative samples $h_i^n$ have been generated as described above for textual and acoustic modalities. 

For each modality, the losses are averaged over each loss element in the batch and used to update the context encoder.
\vspace{-0.1cm}

\begin{equation}
\left\{
\begin{aligned}
\mathcal{L}_{text}^{k} \,& = & L(&H_{text}^{a}, H_{text}^{p}, H_{text}^{n}) \\
\mathcal{L}_{audio}^{k} \, & = & L(&H_{audio}^a, H_{audio}^p, H_{audio}^n) \\
\end{aligned}
\right.
\end{equation}
\begin{equation}
     \mathcal{L}_{contra} = \frac{1}{N}\sum_{k=1}^{N} (
     \mathcal{L}_{text}^{k} + \mathcal{L}_{audio}^{k} )
\label{equ:3}
\end{equation}

where $H_{audio}$ and $H_{text}$ represent acoustic and textual context vectors, respectively, as shown in Figure~\ref{pic:overview}. Minimization of this loss encourages context encoder following: 
\begin{equation}
||h_i^a -h_i^p||_2^2 + m < ||h_i^a -h_i^n||_2^2 , \quad \forall (h_i^a,h_i^p,h_i^n) \in \mathcal{T}
\label{euq:0}
\end{equation}
where $\mathcal{T}$ contains all possible triplets in the training set.

Consequently, the context encoder yields quality representations of underlying context variations which are used later for transferring knowledge to CSS tasks.

\vspace{-0.3cm}
\subsection{Autoregressive Prosodic Modeling}
Inspired by \cite{DBLP:conf/icassp/RenLHZCYZ22}, we adopt an autoregressive prosodic modeling (APM) module to promote fluent and natural prosody. This module ensures that, while generating the current latent prosody vector $LPV_{i+1}$, it takes into account both word-level prosody information and preceding latent prosody vectors ($LPVs$). The APM is constituted by an attention mechanism \cite{DBLP:conf/nips/VaswaniSPUJGKP17} coupled with a pre-trained prosody BERT architecture.
\begin{table*}[t!]
\centering
\caption{\textbf{Subjective evaluation (context-appropriate prosody and naturalness) for different models.}}
\scalebox{0.9}{
    \begin{tabularx}{\textwidth}{X|X|X|X|X|X|X}
    \toprule
    \textbf{Model} & GRU-based & M2CTTS & S1 & S2 & S3 & S4 \\
    \midrule
    \textbf{MOS ($\uparrow$)} & $3.396 \pm 0.107$ & $3.438 \pm 0.104$ & $3.528 \pm 0.097$ & $3.708 \pm 0.108$ & $3.838 \pm 0.110$ & \textbf{$\bm{3.967 \pm 0.120}$}\\
    \bottomrule
    \end{tabularx}
}

\vspace{0.1cm} 

\scalebox{0.88}{
\begin{tabular}{l|c|c|c|c|c|c|c|c}
\toprule
\textbf{Models}& GRU-based vs. M2CTTS & M2CTTS vs. S1 & S1 vs. S2 & S1 vs. S3 & S2 vs. S3 & S3 vs. S4 & GRU-based vs. S4 & M2CTTS vs. S4 \\
\midrule
\textbf{CMOS ($\uparrow$)} & 0.200 & 0.388 & 0.796 & 0.983 & 0.492 & 0.325 & 1.846 & 1.788 \\
\bottomrule
\end{tabular}
}
\label{tab:cmos&mos}
\vspace{-0.45cm}
\end{table*}

\begin{table}[t!]
\centering
\caption{\textbf{Objective evaluation metrics primarily focus on the context-sensitive prosody.} The \texttt{Real} context type uses the correct context for the current synthesized sentence, whereas the \texttt{Fake} type randomly selects from unrelated dialogues.}
\scalebox{0.85}{
\begin{tabular}{c|c|c || c|c|c}
\hline \hline
Method & Set & Type & Mel Loss ($\downarrow$) & Log F0 RMSE ($\downarrow$) & MCD ($\downarrow$) \\
\hline
\multicolumn{2}{c|}{\multirow{2}{*}{GRU-based}} & Real & 3.599  & $0.2949 \pm 0.1192$ & 5.3590\\ 
                       \multicolumn{2}{l|}{}     & Fake & 3.683  & $0.3001 \pm 0.1164$ & 5.3781\\
\hline
\multicolumn{2}{c|}{\multirow{2}{*}{M2CTTS}} & Real & 3.579 & $0.2936 \pm 0.1014$ & 5.3236\\
\multicolumn{2}{l|}{} & Fake & 3.596 & $0.3036 \pm 0.1277$ & 5.3882\\
 
\specialrule{.1em}{.05em}{.05em}
\multirow{8}{*}{CONCSS}& \multirow{2}{*}{S1} & Real & 3.609 & $0.2911 \pm 0.1099$ & 5.3382\\
& & Fake & 3.626 & $0.3203 \pm 0.1093$ & 5.4923\\
\cline{2-6}
\multirow{2}{*}{}& \multirow{2}{*}{S2} & Real & 3.556 & $0.2906 \pm 0.1047$ & 5.2883\\
& & Fake & 3.638 & $0.3311 \pm 0.1417$ & 5.5157\\
\cline{2-6}
\multirow{2}{*}{}&\multirow{2}{*}{S3} & Real & 3.530 & $0.2821 \pm 0.0960$ & 5.2748\\
& & Fake & 3.715  & $0.3272 \pm 0.1455$ & 5.6923\\
\cline{2-6}
\multirow{2}{*}{}&\multirow{2}{*}{S4} & Real & \textbf{3.525} & \textbf{$\bm{0.2803 \pm 0.0961}$} & \textbf{5.2634}\\
&  & Fake & 3.649 & $0.3252 \pm 0.1097$ & 5.6041\\
\hline \hline
\end{tabular}
\label{tab:mse}
}
\vspace{-0.4cm}
\end{table}

 
\begin{table}[]
\centering
\caption{\textbf{Subjective evaluation between different context types.} } 
\scalebox{0.9}{
\begin{tabular}{l|cc|c}
\toprule
\multirow{2}{*}{Model} & \multicolumn{2}{c|}{MOS ($\uparrow$)} & CMOS ($\uparrow$)\\
\cline{2-4}
                       & Real                      & Fake                    & Real vs Fake \\
\hline
GRU-based                     & $3.442 \pm 0.111$       & $3.388 \pm 0.102$       & 0.325      \\
M2CTTS                     & $3.504 \pm 0.100$       & $3.312 \pm 0.112$       & 0.445      \\
\hline
S1                     & $3.638 \pm 0.091$       & $3.250 \pm 0.116$       & 0.492      \\
S2                     & $3.796 \pm 0.076$       & $3.229 \pm 0.101$       & 0.529      \\
S3                     & \textbf{$\mathbf{3.958 \pm 0.074}$}      & $3.308 \pm 0.100$       & \textbf{0.804}      \\
\bottomrule
\end{tabular}
}
\label{tab:fake_real}
\vspace{-0.6cm}
\end{table}
\vspace{-0.25cm}
\section{EXPERIMENTS}
\label{sec:experiments}
\subsection{Experimental Setting}
We conduct experiments on the open-source Chinese conversational speech corpus~\footnote{https://magichub.com/datasets/mandarin-chinese-conversational-speech-corpus-mutiple-devices/} which consists of 10 hours of transcribed Mandarin conversational speech spoken by 30 speakers on certain topics. To better utilize and train the data, we use the ffmpeg toolkit to split continuous dialogues into distinct audio clips, removing non-lexical noises. These clips are then sequenced by utterance order, and their text is converted to phonemes using an open-source tool~\footnote{https://github.com/PaddlePaddle/PaddleSpeech/}. The final processed data totals around 9.2 hours.

\begin{figure}[t]
  \centering
  \includegraphics[width=0.85\linewidth]{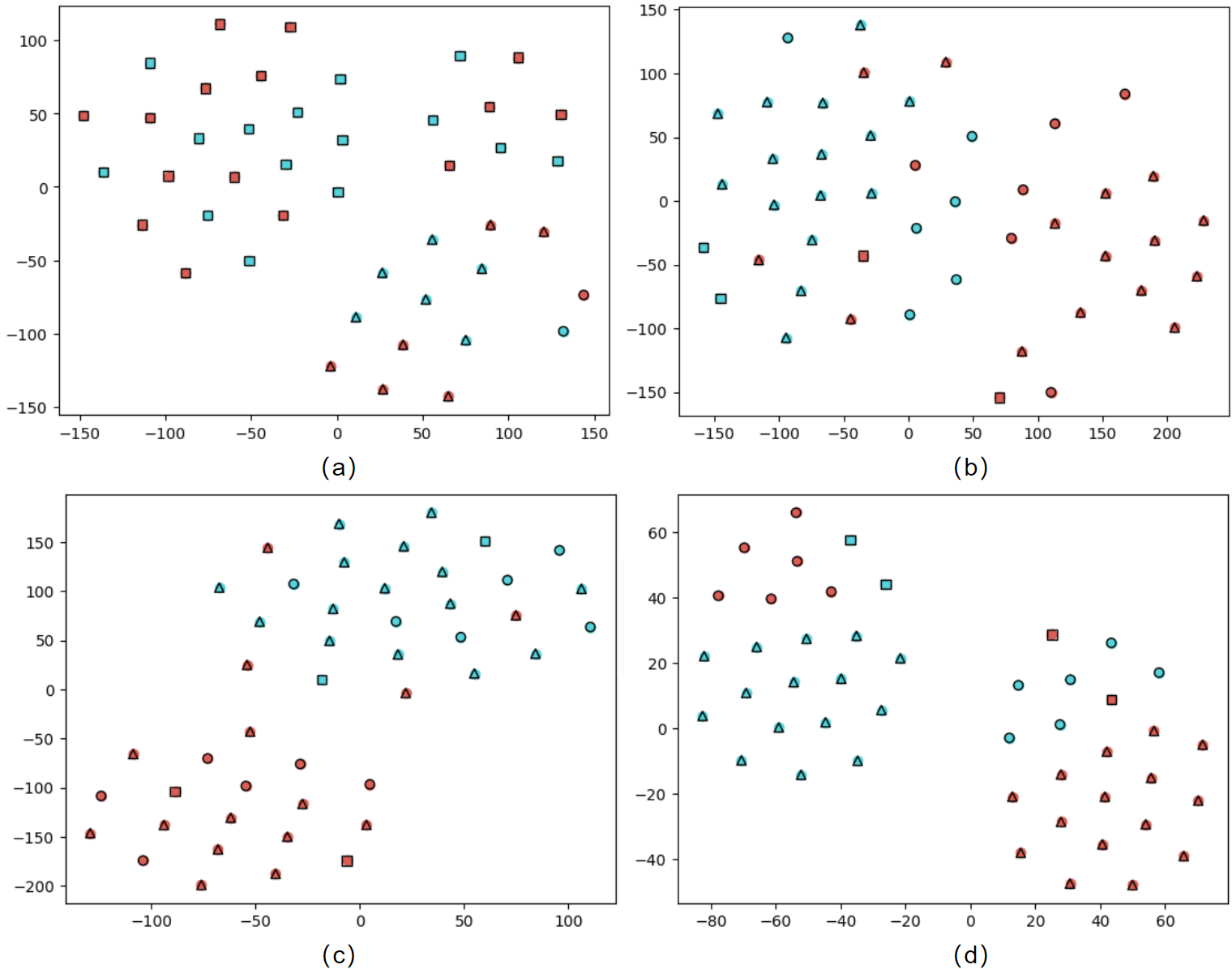}
  \caption{\textbf{T-SNE \cite{van2008visualizing} visualization of context vector distribution from three separate dialogues.} (a) M2CTTS; (b) S1; (c) S2; (d) S3. Identical shapes represent synthesized text from the same dialogue. Red indicates a \texttt{Fake} context, while blue signifies a \texttt{Real} context.} 
  
  \label{pic:tsne}
  \vspace{-0.4cm}
\end{figure}
For the backbone of CSS, we adhere to the vanilla training setups and implementation of VITS. We first pre-train the backbone on the Biaobei Chinese TTS dataset~\footnote{https://www.data-baker.com/open\_source.html} for an initial 5k steps, then the whole CONCSS framework is trained on the Chinese conversation speech corpus for 20k steps with a batch size of 16 to achieve satisfactory results. The Prosody BERT \cite{bert} is finetuned on the prosodic annotation of Biaobei data to learn word-level prosody information.
\vspace{-0.5cm}
\subsection{Compared Models and Evaluation Metrics}
In our experimental evaluation, we conduct comparative and ablation studies on the following seven CSS methods:
\begin{itemize}[topsep=2pt, partopsep=0pt, itemsep=0pt, parsep=0pt, leftmargin=*]
  \item \textbf{GRU-based}: proposed by Guo et al. \cite{guo2021agent} to achieve GRU-based context understanding.
  \item \textbf{M2CTTS}: proposed by Xue et al \cite{xue2023M2CTTS} for multi-modal context understanding (M2CTTS-M6).
  \item \textbf{CONCSS-w/o-APM}: The proposed framework without APM module. With regard to various contrastive learning strategies, we ran ablation studies as follows:
  \begin{itemize}[topsep=0pt, partopsep=0pt, itemsep=1pt, parsep=0pt, leftmargin=*]
    \item  \textbf{S1}: adopts basic contrastive loss proposed by Chopra et al. \cite{DBLP:conf/cvpr/ChopraHL05}. Besides, it does not employ the sampling strategy. 
    \item  \textbf{S2}: utilizes triplet loss \cite{DBLP:conf/cvpr/SchroffKP15} without using the hard negative sampling strategy.
    \item  \textbf{S3}: employs triplet loss and hard negative sampling strategy, as illustrated in Section~\ref{sec:CL}
  \end{itemize}
  \item \textbf{CONCSS}: The proposed CSS framework, denoted as \textbf{S4}.
\end{itemize}

We assess the effectiveness of our proposed method using both subjective and objective measure metrics:
\begin{itemize}[topsep=2pt, partopsep=0pt, itemsep=0pt, parsep=0pt, leftmargin=*]
\item \textbf{Naturalness MOS}: evaluates the overall naturalness of speech on a 1-5 scale, particularly evaluating whether its prosody fits historical dialogues. The greater the score, the better the naturalness.
\item  \textbf{CMOS}: assesses the context-aware prosodic expression. Raters score from -3 (A model is completely better) to 3 (B model is completely better) in 1-point increments. The preference between A and B is denoted as \texttt{(A vs.B)}. A higher score indicates a stronger preference for B.
\item  \textbf{Mel Loss}: calculates the Mean Squared Error (MSE) between the predicted and ground-truth mel-spectrograms.
\item \textbf{Log F0 RMSE and MCD}: are calculated after Dynamic Time Warping \cite{muller2007dynamic} to evaluate prosody-related performance of speech.
\end{itemize}


\vspace{-0.2cm}
\subsection{Comparison of Dialogue-Appropriate Prosody}
We randomly select 20 synthesized speech for each comparison model. Subjective and objective results are compiled and presented in Table~\ref{tab:cmos&mos} and Table~\ref{tab:mse}, and we have the following observations: 1) The MOS score in a multimodal setting surpasses that of a unimodal approach, which can also be supported by previous studies \cite{xue2023M2CTTS,DBLP:conf/icassp/graphCSS}; 2) Within contrastive-based frameworks, S4 achieves optimal performance with a MOS score of $3.967$. Objective metrics also reach the best scores, as highlighted in bold in Table~\ref{tab:mse}. In comparison to baseline systems (GRU-based and M2CTTS), it displays a pronounced preference with CMOS scores of $1.846$ and $1.788$, respectively; 3) Employing triplet loss and hard negative sampling strategy can both enhance prosody performance. Specifically, after utilizing triplet loss, the CMOS score is $0.796$ and the MCD score is $5.2883$. Further incorporation of the negative sample strategy elevates the CMOS to $0.983$ and results in an MCD score of $5.2748$; 4) The APM module also displays an enhancement in prosody, with the MOS score increasing from $3.838$ to $3.967$.

To summarize, both subjective and objective evaluations confirm the efficacy of the contrastive learning approach in enhancing the dialogue-appropriate prosody and naturalness of synthesized speech. Ablation studies are conducted on the contrastive loss function, sampling strategy, and APM module, all of which positively impact the dialogue-appropriate prosody. Compared to basic contrastive loss, the utilization of triplet loss provides more efficient contrastive learning, deepening model's context understanding and thereby facilitating the acquisition of effective context representation. Additionally, Employing the hard negative sampling strategy further boosts contrastive learning, resulting in more dialogue-appropriate prosody. 
\vspace{-0.3cm}
\subsection{Comparison of Context-sensitive Distinctiveness}
To compare discriminability of context vectors generated by different context modeling methods, we first visualize the distance of context vectors. As shown in Fig~\ref{pic:tsne}, we can observe that: 1) non-contrastive framework is locally clustered based on the semantic space of current synthesized sentence (the same shape), which indicates previous frameworks primarily generate context vectors based on the current text, rather than history context; 2) both S2 and S3 methods demonstrate superior context-sensitive discriminative capabilities in comparison to the non-contrastive and S1 methods.

Furthermore, to assess the performance of context-sensitive prosody, we employ a CSS task-specific evaluation method where various CSS frameworks are exposed to diverse contexts. Specifically, given the same synthesized text, we feed compared models with distinct contexts either from the current dialogue or from other irrelevant dialogues, named \texttt{Real} and \texttt{Fake} respectively. Objective and subjective results are shown in Table~\ref{tab:mse} and Table~\ref{tab:fake_real}, respectively. We observe that, in subjective evaluation, S3 has the top CMOS score of $0.804$, indicating that S3 method exhibits the highest sensitivity to context. Besides, the MOS score in S3 shows a clearer distinction between the \texttt{Real} and \texttt{Fake} types by $0.65$, compared to the $0.054$ of the GRU-based method. In objective evaluation, the \texttt{Fake} and \texttt{Real} differences for Mel Loss in the GRU-based system and S3 are $0.084$ and $0.185$ respectively, while for MCD they are $0.0191$ and $0.4175$. We can infer that contrastive-based approaches exhibit greater sensitivity to different contexts than non-contrastive approaches. Both subjective and objective metrics demonstrate that the sampling strategy and triplet loss aid in enhancing the model's context comprehension. This, in turn, results in more discriminative context vectors for diverse context inputs, subsequently influencing the prosodic expression in downstream speech generation. Audio samples are available on the project page~\footnote{https://anonymous.4open.science/w/DEMO-ICASSP2024-5A69/}.
\vspace{-0.3cm}
\section{CONCLUSION}
\label{sec:conclusions}
\vspace{-0.2cm}
In this paper, we introduce CONCSS, a contrastive-based CSS framework that leverages self-supervised training to enhance context understanding. Specifically, we define a pretext task to enable the model to utilize pseudo-labels, thereby increasing context sensitivity to various scenarios. Furthermore, we propose a hard negative sampling strategy to boost context comprehension and the generation of effective context representation. Comprehensive subjective and objective evaluations demonstrate that the proposed method can enhance context comprehension and yield well-representative context vectors, enabling the generation of speech with more appropriate and context-sensitivity prosody. Furthermore, we propose a CSS task-specialized subjective evaluation method to assess models' performance in context understanding and context-sensitive distinctiveness. The effectiveness of the proposed framework is verified in comprehensive experiments.
\vspace{-0.3cm}
\section{Acknowledgements}
\label{sec:ack}
\vspace{-0.1cm}
The work was supported by the National Natural Science Foundation of China (NSFC) (No.62271083), the Fundamental Research Funds for the Central Universities (No.2023RC13), the open research fund of The State Key Laboratory of Multimodal Artificial Intelligence Systems (No.202200042).




\vfill\pagebreak
\bibliographystyle{IEEEbib}
\bibliography{Template}

\end{document}